\documentclass[lettersize,journal]{IEEEtran}
\usepackage{amsmath,amsfonts}
\usepackage{algorithmic}
\usepackage{algorithm}
\usepackage{array}
\usepackage[caption=false,font=normalsize,labelfont=sf,textfont=sf]{subfig}
\usepackage{textcomp}
\usepackage{stfloats}
\usepackage{url}
\usepackage{verbatim}
\usepackage{graphicx}
\usepackage{caption}
\usepackage{hyperref}
\usepackage[most]{tcolorbox}
\usepackage{xcolor}

\begin{document}

\title{SMART SLM: Structured Memory and Reasoning Transformer, A Small Language Model for Accurate Document Assistance}

\author{{Divij Dudeja, Mayukha Pal, Senior Member, IEEE}
 

\thanks{(Corresponding author: Mayukha Pal)}
\thanks{Mr. Divij Dudeja is a Data Science Research Intern at ABB Ability Innovation Center, Hyderabad 500084, India and also a B.Tech undergraduate student from Department of Computer Science Engineering, Indian Institute of Information Technology, Nagpur, Nagpur 441108, IN.}

\thanks{Dr. Mayukha Pal is with ABB Ability Innovation Center, Hyderabad-500084, IN, working as Global R\&D Leader – Cloud \& Advanced Analytics (e-mail: mayukha.pal@in.abb.com).}
}

\maketitle

\begin{abstract}

The user of Engineering Manuals (EM) finds it difficult to read EM's because they are long, have a dense format which includes written documents, step by step procedures, and standard parameter lists for engineering equipment.Off-the-shelf transformers, especially compact ones, treat this material as a flat stream of tokens. This approach leads to confident but incorrect numeric answers and forces the models to memorize separate facts inefficiently. SMART (Structured Memory and Reasoning Transformer) offers a different and practical solution to the above problem. SMART structures its processing by using a hierarchical approach, and is based upon three main job categories: (1) A syntax-aware Fact Extractor (Grammarian) Tree-LSTM which extracts facts as subject-relation-object relations from EM sentences; (2) A compact indexed memory MANN (Memory Augmented Neural Network) that indexes these Rational Subject-Relation-Objects as 384-dimensional vectors that are associated with the source of the information, and (3) A 6-layer Transformer that learns to fuse the previously retrieved facts into its generated response. The entire SMART model utilizes 45.51M parameters, which is 64\% less than GPT-2 (124M) and  69\% less than BERT (133M), and it achieves a 21.3\% higher accuracy than GPT-2, indicating that SMART fits the data better with the least amount of processing requirements.

SMART employs dual modes of inference: an indexed fast path for known documents (sub-second answer times) and an indexed dynamic path assisted by RAGs for new uploads (FAISS Top-20 results with memory severed at 64 slots). In real world deployment, this framework leads to more well-supported results with reduced hallucinations than comparable small transformer models. SMART presents an effective recipe with regard to trustworthy document-based assistance, namely extract good-quality assertions, provide checks within documents, and employ an efficient dispatcher with these assertions during answering.

\end{abstract}

\section{INTRODUCTION}

Transformers have reshaped the way machines understand and generate language. Using attention mechanisms to weigh the importance of different words, transformer-based models can capture long-range relationships in text and produce fluent, context-aware responses\cite{vaswani2017attention}. This breakthrough underpins many modern NLP (natural language processing) systems used for tasks such as translation, summarization, and question answering. At the same time, there is growing interest in compact models with modest parameter counts that can run efficiently on limited hardware while still providing useful, reliable behavior.

Technical documents and device manuals present a particular challenge for language models. These documents are made up of description, step-by-step instructions and multi-column, tabular-format tables that contain numerical values and settings which are described with precision. The model must be able to read fluently in English and retrieve from the records information that has been recorded and is factually correct. In addition, when using off-the-shelf transformer models, the small number of words in any order on the page will make it possible for the model to produce a reasonable response with a high probability of being inaccurate. Therefore, creating a model that can accurately respond and be of adequate size and structure is the fundamental issue we are addressing in this study.

\subsection{Motivation and Problem Statement}
Engineers and technical personnel rely extensively on quick, accurate answers from large documents. However, the current approaches for determining appropriate answers in SLMs struggle to provide accurate answers from these types of documents without causing errors. Some of the major challenges include:

\textbf{Structured Content: } Unlike flat text formats, manuals often contain tables and parameter lists that define relationships. The loss of this structure is a huge detractor in correctly communicating the comprehensive detail within a manual.

\textbf{Factual Accuracy: }  Users expect numeric values and recommendations to be unambiguous. A mistake of only one unit could be destructive in practice. Therefore, the SLM must provide facts that can be verified (i.e. not "hallucinated").

\textbf{Limited Memory: } A small language model lacks sufficient memory capacity to store thousands of discrete facts and retain proficiency in the language. A balanced approach requires separation of memorized facts and generated sentences.

\textbf{Latency and Usability: } To fully assess and analyze a complete manual is computationally intensive and therefore produces excessive delays. On the other hand, not having the complete context available can lead to errors. A functional SLM must use both speed and accuracy to deliver quality results.

\subsection{Our Approach}
We address these challenges by dividing the task into two complementary functions: extracting high-quality, structured facts from documents, and using a compact reasoning engine that consults those facts when producing answers. This separation keeps the language model small and focused on fluent output while delegating factual precision to an external memory.

Our approach is fundamentally different from existing methods in several key aspects:

\textbf{Grammarian-first fact extraction:} We use a Tree-LSTM (Long Short Term Memory) based FactExtractor to parse sentence structure and convert complex sentences and table rows into canonical (subject(s), relation(r), object(o)) facts. This step produces semantically clear high-quality facts rather than noisy text fragments.

\textbf{Librarian-style external memory:} Extracted facts are stored in a Memory-Augmented Neural Network (MANN). Facts are represented as compact vectors, indexed, and made available for exact retrieval. Memory acts as a stable source of truth that the language engine can query.

\textbf{Gated memory fusion in the encoder:} Instead of letting the transformer attempt to memorize facts, the encoder fuses memory-derived evidence into token representations via a gated attention mechanism. This yields responses that are fluent and fact-based.

\textbf{Dual-mode inference:} For known documents we load pre-computed memory matrices for instant response. For new documents, we first perform lightweight retrieval (RAG) to find the most relevant chunks and then run fact extraction only on that subset, reducing latency while preserving flexibility.

\textbf{Provenance and numeric normalization:} Facts retain links to their source snippets and include normalized numeric representations (units, ranges). This improves traceability and reduces errors that arise from unit mismatches.

\subsection{Contributions}
This paper makes the following key contributions:
\begin{itemize}
    \item A novel Grammarian→Librarian→Transformer architecture (SMART) that combines a Tree-LSTM fact extractor with memory-augmented storage and a compact transformer reasoning engine, designed specifically for technical-document question answering.
    \item A gated memory-fusion encoder layer that integrates retrieved facts into token-level representations, enabling small models to generate fluent answers while relying on an external factual store.
    \item A practical dual inference strategy that supports both pre-indexed fast lookup for known documents and RAG-assisted dynamic compilation for new documents   balancing latency, flexibility, and accuracy.
    \item An end-to-end implementation and evaluation framework that emphasizes factual precision, provenance, and usability for domain experts, together with ablation studies that quantify the benefits of each component.
    
\end{itemize}

\section{PRIOR ART}

\subsection{Transformer, long-context models and SLM's}

Transformers with self-attention power modern NLP and can be extended for long contexts using sparse or sliding-window attention, hierarchical encodings, and alternative sequence operators. These methods let models process thousands of tokens end-to-end, but they still view documents as a flat token stream, which can obscure structured content such as tables and parameter lists.

SLM's are compact variants created by distillation, parameter sharing, pruning, and quantization (examples: DistilBERT, TinyBERT, MobileBERT)\cite{sanh2019distilbert}\cite{jiao2020tinybert}\cite{sun2020mobilebert}. They run efficiently on limited hardware but trade capacity for size: reduced memorization makes precise factual recall harder, and compressed models are more prone to subtle factual errors when asked for domain-specific numeric values\cite{reddy2025contextual}.

\subsection{Retrieval and RAG-style systems}
RAG (Retrieval-Augmented Generation) and related pipelines combine a fast retriever that finds relevant passages with a generator that conditions on those passages. Dense embedding based retrievers (paired with approximate nearest-neighbor search) make retrieval fast and scalable; cross-encoder rerankers can improve precision. RAG-style systems are powerful for open-domain question answering because they reduce the need for the generator to memorize facts. However, typical RAG pipelines still rely on the generator to interpret retrieved text and can inherit errors from noisy passages; they also do not guarantee compact, canonical fact representations that are directly queryable\cite{lewis2020retrieval}\cite{tiwari2025ontorag}.

\subsection{Memory-augmented neural networks}
External-memory architectures give a model explicit storage that can be read from and written to during processing\cite{santoro2016meta}. Earlier work explored differentiable memory banks that allow models to store and retrieve discrete pieces of information; more recent approaches show that augmenting a language model with an external, indexed memory can materially reduce hallucinations and increase factual consistency. Nevertheless, the utility of an external memory depends on the quality of the content stored there: raw text snippets are convenient but noisy, and naive memory population strategies can waste capacity on redundant or low-value facts.

\subsection{Tree-structured and syntactic models for extraction}
Models that operate on dependency or constituency trees such as Tree-LSTM variants are explicitly designed to capture the syntactic and logical relationships inside sentences\cite{tai2015improved}. This makes them well suited to extracting relations and structured facts from complex, nested sentences that occur frequently in technical prose. Prior systems have used tree-based encoders for information extraction and for producing compact semantic representations, but they are rarely paired directly with an external memory and a small transformer generator in a tightly integrated pipeline for document QA.

\section{METHODOLOGY}

This section explains how SMART works in an easy-to-follow way. First, we give a brief overview of the whole pipeline and then detail each component: data preparation and retrieval, fact extraction (the “Grammarian”), memory storage (the “Librarian”), the transformer reasoning engine and its gated memory fusion, training and loss functions, and the two inference modes that we use in practice.

\subsection{Smart System architecture overview}

SMART is built as three cooperating parts:

Grammarian (FactExtractor) reads short passages and converts sentences into compact facts of the form (subject, relation, object). Each fact is also connected to the exact passage and location from which it came from (provenance).

Librarian (Memory). Stores the extracted facts as fixed-length numeric rows so they can be quickly searched and retrieved.

Reasoning \& Generation (SMART Transformer). A small transformer model that consults the Librarian during its understanding step and then writes a short, natural-language answer.

Two inference modes are supported:

Pre-indexed mode (Path A): documents processed offline → instant answers.
-
RAG-assisted dynamic mode (Path B): for new documents, we first retrieve likely passages and then run fact extraction only on those passages to build a small, focused memory, which keeps latency manageable.

\begin{figure}[!htbp]
    \centering
    \includegraphics[width=0.79\linewidth]{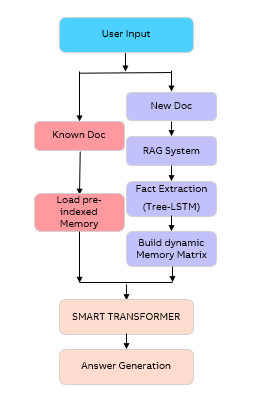}
    \caption{SMART architecture}
    \label{fig:placeholder}
\end{figure}

\begin{figure}[!htbp]
    \centering
    \includegraphics[width=0.8\linewidth]{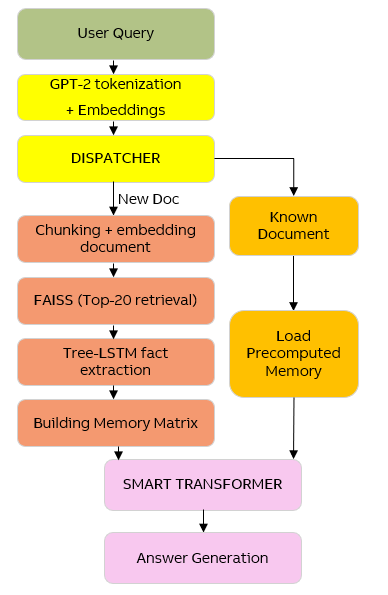}
    \caption{Query Processing Pipeline}
    \label{fig:system}
\end{figure}

\subsection{Data Preparation and retrieval}

What we store and why. Documents are first divided into manageable passages so that the system can quickly find the relevant text. The index contains 380,438 passages (each ~100 tokens). The model was trained with additional small-English data (TinyStories) during the early stages to learn fluent English\cite{eldan2023tinystories}.

Chunking. Long documents are split into chunks of ~150 words with ~30\% overlap. Overlap helps to avoid cutting a useful sentence in half.

Passage embedding and search. Each chunk is converted to a numeric vector using a small embedding model (all-MiniLM-L6-v2). Vectors are normalized to unit length and indexed with FAISS IndexFlatIP. For a user query, we embed the query and retrieve the Top-20 nearest passages. These passages feed the next step.

\subsection{Grammarian   extracting canonical facts}

The goal is to turn each useful sentence into a small structured fact that is easy to store and retrieve.

\subsubsection{Finding candidate spans}
\begin{itemize}
    \item A syntactic parser (spaCy + benepar) identifies sentence structure.

    \item Heuristics pick likely subject, relation, and object spans using standard dependency markers (nsubj, dobj, etc.). If heuristics fail, a fallback chooses the nearest noun phrase and verb.
\end{itemize}

\subsubsection{How the Tree-LSTM summarizes a phrase}
We use a Tree-LSTM to compute a vector for each phrase (subject/relation/object). A Tree-LSTM merges information from a phrase’s child parts (words and subphrases) into a single vector.

Mathematically, for a node \( j\) with children \( C(j)\):
\begin{equation}i_j \&= \sigma\big(W^{(i)} x_j + U^{(i)} \tilde{h}*j + b^{(i)}\big) \end{equation}
\begin{equation}f*{jk} \&= \sigma\big(W^{(f)} x_j + U^{(f)} h_k + b^{(f)}\big) \quad \forall k\in C(j)\end{equation}
\begin{equation}o_j \&= \sigma\big(W^{(o)} x_j + U^{(o)} \tilde{h}*j + b^{(o)}\big) \end{equation}
\begin{equation}u_j \&= \tanh\big(W^{(u)} x_j + U^{(u)} \tilde{h}*j + b^{(u)}\big)\end{equation}
\begin{equation}c_j \&= i_j \odot u_j + \sum*{k\in C(j)} f*{jk} \odot c_k \end{equation}
\begin{equation}h_j \&= o_j \odot \tanh(c_j)\end{equation}

\subsubsection{Producing the fact vectors}

For each chosen span (subject, relation, object) we apply a small projection:
\begin{equation}
v_s = \text{GELU}(W_s h_{\text{span}} + b_s) \quad\text{(128 dims)}
\end{equation}

Similarly for \((v_r)\) and \((v_o)\). The final memory row for the fact is:

\begin{equation}
m = [v_s ,|, v_r ,|, v_o] \in \mathbb{R}^{384}.
\end{equation}

\subsection{Librarian   storing and retrieving facts}
The Librarian is more than a passive file of facts   it is an organized, searchable memory that the Transformer consults when answering a question. In straightforward terms, the Librarian stores many compact fact vectors (one per extracted fact), finds the small set that is most relevant to a question, and presents those facts to the model in a form the model can use easily. We implement the Librarian as a hybrid MANN: a learned, vectorized memory representation (the “neural” part) combined with a fast, disk-backed index (FAISS) for large-scale retrieval.

\subsubsection{What the memory contains and how it is stored}
\begin{itemize}
    \item Memory rows: each extracted fact is stored as a fixed-length numeric row \((m_i \in \mathbb{R}^{384})\). This row is the concatenation of three 128-dim projections: subject \((v_s)\), relation \((v_r)\), object \((v_o)\).
    \item Metadata/provenance: for each row we also keep human-readable fields(span text, passage id, doc id, sentence id, character offsets).
    \item Indexing: the vectors are L2-normalized and stored in a FAISS IndexFlatIP. Using normalized vectors means inner product search approximates cosine similarity, which is a robust measure of semantic closeness.

\end{itemize}

\subsubsection{How the Librarian answers the question of relevance}
\begin{itemize}
    \item Fast coarse retrieval (FAISS):\begin{itemize}
        \item We embed the user query into a 384-dim vector q (mean-pooled token embedding) and normalize it.
        \item We run a Top-20 search in FAISS to return the most semantically similar passages. This step is fast even for very large collections because FAISS is optimized for nearest-neighbor search.
    \end{itemize}
    \item Slot collection and content re-scoring: \begin{itemize}
        \item For each retrieved passage, we obtain up to 4 extracted slots (facts). Collectively these produce a candidate memory matrix 
        \begin{equation}
            M = [m_1; m_2; \dots; m_N] \in \mathbb{R}^{N\times 384}
        \end{equation}
        where typically \((N \le 80)\) (20 passages × up to 4 slots). We then optionally trim (M) to the top 64 rows by confidence or by inner product with (q). This yields the small, focused memory the encoder will consult.
    \end{itemize}
\end{itemize}

\subsubsection{How the model reads memory}
Once a compact memory matrix (M) is assembled, the encoder computes a weighted read vector that summarizes the relevant facts. A single-layer summary (as used in SMART) is:

\begin{equation}
    \tilde{q} = q W_Q^{(m)} \in \mathbb{R}^{d_k}\
\end{equation}
\begin{equation}
\tilde{K} = M W_K^{(m)} \in \mathbb{R}^{N\times d_k}\
\end{equation}
\begin{equation}
    \tilde{V} = M W_V^{(m)} \in \mathbb{R}^{N\times d_v}\
\end{equation}
\begin{equation}
    \alpha = \text{softmax}\left(\frac{\tilde{q},\tilde{K}^\top}{\sqrt{d_k}}\right)\in\mathbb{R}^N\
\end{equation}
\begin{equation}
    c_{\text{mem}} = \alpha^\top \tilde{V}\in\mathbb{R}^{d_v}.
\end{equation}
\begin{itemize}
    \item \((\alpha)\) is a probability vector that assigns higher weight to memory rows more relevant to the query.
    \item \((c_{\text{mem}})\) is the single context vector that summarizes the retrieved facts for that encoder layer.

\end{itemize}

\subsection{The SMART Model Architecture}
We now explain the language model and how it consults the memory. Start with the standard building blocks and then the added memory wiring.

\subsubsection{Model Parameters}
We list the exact model sizes and other numeric choices used in SMART. These values were chosen to balance compactness, speed, and sufficient capacity for factual reasoning.

\begin{equation}
   \text{Model hidden dimension: } (d_m): 384
\end{equation}
\begin{equation}
   \text{Feed-forward inner dimension: } (d_f): 1536
\end{equation}
\begin{equation}
   \text{Number of transformer layers: } ( L_\text{transformer}): 6
\end{equation}
\begin{equation}
   \text{Number of attention heads: } (H): 8
\end{equation}
\begin{equation}
   \text{Per-head key/query dimension: } (d_k): 48
\end{equation}
\begin{equation}
   \text{Tree-LSTM token vector dimension: } 384
\end{equation}
\begin{equation}
   \text{Memory vector: } (d_memory): 384
\end{equation}
\begin{equation}
   \text{Vocabulary Size: } 50257
\end{equation}
\begin{equation}
   \text{Total Model Parameters: } (\theta): 45.51M
   \end{equation}

\subsubsection{Token embeddings and positional embeddings}
\mbox{}\\
    Token embeddings:

Each input word piece (token) is mapped to a learned vector called a token embedding. If the token index is (i), its embedding is \((E[i]\in\mathbb{R}^{384})\). When a query phrase is tokenized into (T) tokens, we form the token embedding matrix:
\begin{equation}
    X_{0} = [E[t_1]; E[t_2]; \dots; E[t_T]] \in \mathbb{R}^{T\times 384}.
\end{equation}

Positional embeddings:

Transformers are order-agnostic by default; we therefore add a learned positional embedding for each token position. Let \((P[p]\in\mathbb{R}^{384})\) be the learned vector for position (p). The input to the encoder is:
\begin{equation}
    X_{\text{input}} = X_{0} + P[0{:}T-1] \in \mathbb{R}^{T\times 384}.
\end{equation}

The positional encoding follows the sinusoidal pattern:

\begin{equation}
    \text{PE}(i, 2k) = \sin\left( \frac{i}{10000^{2k/d}} \right)
\end{equation}
\begin{equation}
    \text{PE}(i, 2k+1) = \cos\left( \frac{i}{10000^{2k/d}} \right)
\end{equation}

\subsubsection{Scaled dot-product attention (single head)}
For a single attention head we compute three linear projections of the input token vectors (X):
\begin{equation}
    \text{Queries: } (Q = XW_Q)
\end{equation}
\begin{equation}
    \text{Keys: } (K = XW_K)
\end{equation}
\begin{equation}
    \text{Values: } (V = XW_V)
\end{equation}
where \((W_Q, W_K, W_V \in \mathbb{R}^{384\times d_k})\) and \((d_k=48)\).

Scaled dot-product attention is:

\begin{equation}
    \text{Attention}(Q,K,V)=\text{softmax}\left(\frac{QK^\top}{\sqrt{d_k}}\right)V.
\end{equation}

\subsubsection{Multi-head Attention}
We run (H=8) heads in parallel. For head (h) we compute:
\begin{equation}
    \text{head}_h = \text{Attention}(QW_Q^{(h)},,KW_K^{(h)},,VW_V^{(h)}).
\end{equation}
All head outputs are concatenated and linearly projected back to model dimension:
\begin{equation}
    \text{MHA}(X) = \big[ \text{head}_1, \dots, \text{head}_8 \big] W_O,
\end{equation}

\subsubsection{Feed-forward network}
After attention, each token vector passes through a small two-layer network applied tokenwise:
\begin{equation}
    \text{FFN}(x) = W_2,\text{GELU}(W_1 x + b_1) + b_2,
\end{equation}

\subsubsection{Memory-attention and gated fusion}

the transformer layer does not only attend to tokens; it also reads a small external memory of fact vectors and fuses that information into token representations.
Memory-attention (per transformer layer) Given:
\begin{itemize}
    \item memory matrix \((M \in \mathbb{R}^{N\times 384})\) (N memory rows, each 384-d)
    \item a query intent vector \((q \in \mathbb{R}^{384})\) representing the user query
\end{itemize}

we compute a compact memory context vector \((c_{\text{mem}})\) as follows
\begin{equation}
    \tilde{q} = q W_Q^{(m)} \in \mathbb{R}^{d_k},
\end{equation}
\begin{equation}
    \tilde{K} = M W_K^{(m)} \in \mathbb{R}^{N\times d_k},
\end{equation}
\begin{equation}
    \tilde{V} = M W_V^{(m)} \in \mathbb{R}^{N\times d_v}.
\end{equation}
Compute attention weights and context:

\begin{equation}
    \alpha = \text{softmax}\left(\frac{\tilde{q},\tilde{K}^\top}{\sqrt{d_k}}\right) \in \mathbb{R}^{N},   
c_{\text{mem}} = \alpha ,\tilde{V} \in \mathbb{R}^{d_v}.
\end{equation}
In our implementation we align sizes so \((d_v = d_k \cdot H = 384)\); hence \((c_{\text{mem}}\in\mathbb{R}^{384})\).

Gated fusion: 

Let \((X_{\text{self}} \in \mathbb{R}^{T\times384})\) be the token-level output from self-attention in the same layer. We fuse the memory vector into token vectors using a learned scalar gate per encoder block.

Let \((\gamma^{(b)})\) be a learned scalar for block (b); define the gate
\begin{equation}
    g^{(b)} = \sigma(\gamma^{(b)}),\qquad g^{(b)}\in(0,1).
\end{equation}
Broadcast \((c_{\text{mem}})\) to token length (T) as \((\widetilde{c}_{\text{mem}}\in\mathbb{R}^{T\times384})\). The fused token representation is:
\begin{equation}
    X_{\text{fused}} = g^{(b)} \odot X_{\text{self}} + (1-g^{(b)}) \odot \widetilde{c}_{\text{mem}}.
\end{equation}

\subsubsection{Complete transformer block}
The complete transformer block used by SMART is a paired unit that (a) builds a memory-informed representation of the user query and (b) produces tokens auto regressively.
\mbox{}\\
Each encoder layer executes the following steps :
\begin{equation}
    X_{\text{self}} = \text{MHA}(X_{\text{in}})),(X_{\text{in}}\in\mathbb{R}^{T\times384}.
\end{equation}
\begin{equation}
    X_{\text{fused}} = g^{(b)} X_{\text{self}} + (1-g^{(b)}) \widetilde{c}_{\text{mem}}
\end{equation}
\begin{equation}
    X' = \text{LayerNorm}(X_{\text{in}} + X_{\text{fused}})
\end{equation}
\begin{equation}
    (X_{\text{ffn}} = \text{FFN}(X'))
\end{equation}
\begin{equation}
    X_{\text{out}} = \text{LayerNorm}(X' + X_{\text{ffn}})
\end{equation}

Each decoder layer executes the following steps :
\begin{equation}
    Q_s = Y_{\text{in}}W_Q^{(s)}\in\mathbb{R}^{U\times d_k},\quad
\end{equation}
\begin{equation}
    K_s = Y_{\text{in}}W_K^{(s)}\in\mathbb{R}^{U\times d_k},\quad
\end{equation}
\begin{equation}
    V_s = Y_{\text{in}}W_V^{(s)}\in\mathbb{R}^{U\times d_k},\
\end{equation}
\begin{equation}
    Y_{\text{self}}  \text{softmax}\Big(\frac{Q_sK_s^\top}{\sqrt{d_k}} + \text{Mask}_{\text{causal}}\Big)V_s \in\mathbb{R}^{U\times384}
\end{equation}
\begin{equation}
    Y' = \text{LayerNorm}(Y_{\text{in}} + Y_{\text{self}})
\end{equation}
\begin{equation}
    Q_c = Y'W_Q^{(c)} \in\mathbb{R}^{U\times d_k}
\end{equation}
\begin{equation}
    K_e = ZW_K^{(e)} \in\mathbb{R}^{T\times d_k}
\end{equation}
\begin{equation}
    V_e = ZW_V^{(e)} \in\mathbb{R}^{T\times d_v}
\end{equation}
\begin{equation}
    Y_{\text{cross}} = \text{softmax}\left(\frac{Q_c K_e^\top}{\sqrt{d_k}}\right)V_e \in\mathbb{R}^{U\times384}.
\end{equation}
\begin{equation}
    Y'' = \text{LayerNorm}(Y' + Y_{\text{cross}})
\end{equation}
\begin{equation}
    Y_{\text{ffn}} = \text{FFN}(Y'')= W_2;\text{GELU}(W_1 Y''+b_1) + b_2\in\mathbb{R}^{U\times384}
\end{equation}
\begin{equation}
    Y_{\text{out}} = \text{LayerNorm}(Y'' + Y_{\text{ffn}}) \in\mathbb{R}^{U\times384}
\end{equation}

\subsection{Training Strategy}

We train SMART in stages to stabilize learning and to teach memory how to align with queries without harming language fluency.
\subsubsection{Stage 1: Language Pretraining }
\mbox{}\\
Objective: next-token prediction (cross-entropy),with using the context window of 4
Key hyperparameters: lr = 4e-5, warmup = 3000 steps, batch size = 32, total steps for the checkpoint used = 240,000 steps.
\subsubsection{Stage 2: memory warmup}
\mbox{}\\
Objective: align query projections and memory rows using mean-squared error (MSE).
Loss:
\begin{equation}
    \mathcal{L}*{\text{MSE}} = |q*{\text{proj}} - m|_2^2.
\end{equation}
Optimizer: AdamW, lr = 1e-4, weight decay = 0.01, gradient clipping = 1.0

\subsubsection{Stage 3: joint fine-tuning}
\mbox{}\\
Objectives:
\begin{itemize}
    \item Contrastive retrieval loss (InfoNCE): pull correct memory rows toward queries and push negatives away.
    \begin{equation}
        \mathcal{L}*{\text{InfoNCE}} = -\log\frac{\exp(q\cdot m^+/ \tau)}{\sum*{j}\exp(q\cdot m_j / \tau)},\quad \tau=0.07
    \end{equation}
    We use in-batch negatives plus sampled negatives (~31 negatives per positive)
    \item Optional reconstruction loss: decoder reconstructs the textual triple from (m) with cross-entropy; weight ~0.5
\end{itemize}
Combined Loss:
\begin{equation}
    \mathcal{L} = 1.0\cdot\mathcal{L}*{\text{MSE}} + 1.0\cdot\mathcal{L}*{\text{InfoNCE}} + 0.5\cdot\mathcal{L}_{\text{recon}}
\end{equation}

Optimizer and settings: AdamW, lr = 1e-4, batch size = 32, mixed precision disabled for stability

\section{RESULTS AND ANALYSIS}
\subsection{Main Performance Results}

Table I presents the comprehensive comparision of our SMART model against all baseline architectures:

\mbox{}\\

\begin{table}[htbp]
\centering
\caption{ Performance Comparison: SMART vs Transformer Baselines}
\label{tab:cgt_comparison}

\renewcommand{\arraystretch}{1.25} 

\begin{tabular}{|l|c|c|}
\hline
\textbf{Model} & \textbf{Parameters (M)} & \textbf{Final Loss} \\
\hline
DistilBERT & 89.8 & 10.430 \\
GPT-2 & 124.4 & 2.787 \\
BERT & 133.0 & 10.460 \\
Pure Transformer & 52.0 & 3.456 \\
\hline
\textbf{SMART (Our Model)} & \textbf{45.51} & \textbf{2.341} \\
\hline
Improvement vs GPT-2 & -63.4\% & +16.0\% \\
Improvement vs Pure Transformer & -12.5\% & +32.3\% \\
\hline
\end{tabular}
\end{table}

\subsection{Detailed BELU and ROUGE Evaluation}

Table II presents the comprehensive evaluation of our model with the pure transformer model, on various metrics:

\begin{table}[htbp]
\centering
\caption{ Performance Comparison: SMART vs  Pure Transformer Model }
\label{tab:cgt_comparison}

\renewcommand{\arraystretch}{1.25} 

\begin{tabular}{|l|c|c|}
\hline
\textbf{Metrics} & \textbf{SMART (SLM) } & \textbf{Pure Transformer} \\
\hline
BLEU-1 Score & 0.1445 &  0.0238 \\
BLEU-2 Score & 0.0512 & 0.0080 \\
BLEU-4 Score & 0.0148 & 0.0038 \\
\hline
ROUGE-1 Score & 0.2032 &  0.0511 \\
ROUGE-2 Score & 0.0394 & 0.0015 \\
ROUGE-L Score & 0.1734 & 0.0481 \\
\hline
Response Time & 0.4578 & 0.2696 \\
\hline
\end{tabular}
\end{table}

The results shows remarkable improvement across all evaluation metrics:
 
\begin{itemize}
    \item \textbf{BLEU Scores: } 458\%-587\% improvement indicates superior n-gram overlap 
    \item \textbf{ROUGUE Scores: } Upto 2438\% improvement in recall oriented metrics  
    \item \textbf{Efficiency Trade-off: } 68\% longer response time than pure transformer, for quality results 
\end{itemize}

\subsection{Performance Visualization}
\begin{figure}[!htbp]
    \centering
    \includegraphics[width=1.05\linewidth]{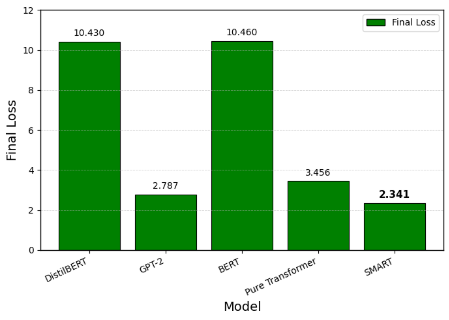}
    \caption{ Final loss of various models comparison}
\end{figure}
The Figure 3 compares the final loss of five models on the same task; lower is better. Our model, SMART, has the lowest final loss at 2.341, noticeably below GPT-2 (2.787) and Pure Transformer (3.456)\cite{radford2019language}. This represents about a 16.0\% reduction in loss versus GPT-2 and roughly a 32.3\% reduction versus the Pure Transformer, indicating SMART achieves better fit on the trained objective. Because SMART also uses far fewer parameters (45.51M vs GPT-2’s 124.4M), the plot highlights that SMART is both more accurate and more parameter-efficient than those baselines.

\subsection{Parameter Efficiency Analysis}
To quantify parameter efficiency, we define the efficiency
metric:
\begin{equation}
    \text{Efficiency}(\text{model}) = \frac{1}{\text{Parameters} \times \text{Loss}}
\end{equation}

\begin{table}[htbp]
\centering
\caption{\textbf{Parameter Efficiency Analysis}}
\label{tab:parameter_efficiency}

\renewcommand{\arraystretch}{1.25}

\begin{tabular}{|l|c|c|c|}
\hline
\textbf{Model} & \textbf{Parameters (M)} & \textbf{Loss} & \textbf{Efficiency Score} \\
\hline
DistilBERT & 89.8 & 10.430 & $1.07 \times 10^{-3}$ \\
GPT-2 & 124.4 & 2.787 & $2.88 \times 10^{-3}$ \\
BERT & 133.0 & 10.460 & $7.19 \times 10^{-4}$ \\
Pure Transformer & 52.0 & 3.456 & $5.56 \times 10^{-3}$ \\
\hline
\textbf{SMART (Ours)} & \textbf{45.51} & \textbf{2.341} & $\mathbf{9.12 \times 10^{-3}}$ \\
\hline
\textbf{Improvement vs Best} & \textbf{-12.5\%} & \textbf{+32.3\%} & \textbf{+117.3\%} \\
\hline
\end{tabular}
\end{table}

\textbf{ Our model SMART SLM achieves 117\% improvement in parameter efficiency compared to Pure Transformer Baseline.}

\subsection{Qualitative Analysis: Generated Responses}
To demonstrate the practical effectiveness of our approach, we present actual generated responses from our SMART model with highlighted formatting

\begin{tcolorbox}[
  colback=cyan!10,
  colframe=cyan!60,
  boxrule=1.2pt,
  arc=2pt,
  left=6pt,
  right=6pt,
  top=6pt,
  bottom=6pt
]
\textbf{Example 1:}

\textit{Question:} How do I verify a protection setting before testing in the field?

\textbf{SMART Response:} Check the device type and software version first so you know it matches the document. Save the current settings so nothing is lost. Read the related section once to understand what the setting is meant to do. Try it in a safe test setup if possible. While testing, keep watching the device messages and alarms.
\end{tcolorbox}

\vspace{0.8em}

\begin{tcolorbox}[
  colback=green!10,
  colframe=green!70,
  boxrule=1.2pt,
  arc=2pt,
  left=6pt,
  right=6pt,
  top=6pt,
  bottom=6pt
]
\textbf{Example 2:}

\textit{Question:} What’s a safe procedure for testing a new protection curve on site?

\textbf{SMART Response:} First make sure the area is safe and people are informed. Use test equipment instead of the real system load. Apply the test slowly and see how the device reacts. If something looks wrong, stop the test. After finishing, put back the original settings and note what happened.
\end{tcolorbox}

\vspace{0.8em}

\begin{tcolorbox}[
  colback=purple!10,
  colframe=purple!60,
  boxrule=1.2pt,
  arc=2pt,
  left=6pt,
  right=6pt,
  top=6pt,
  bottom=6pt
]
\textbf{Example 3:}

\textit{Question:} What should I include when documenting a protective device configuration?

\textbf{SMART Response:} Write down the device name, version, and basic settings used. Add simple diagrams or notes on connections. Mention why the settings were chosen and any tests done. Keep the language clear so someone else can understand it later.
\end{tcolorbox}

\section{DISCUSSION}

\subsection{Strengths and Advantages of the SMART}
\begin{itemize}
    \item \textbf{Factual reliability: } Facts stored in the Librarian memory are discrete or canonical in nature and have been derived using source passages. When the model retrieves an answer, it consults the available evidence  rather than attempting to remember everything based on weighing factors alone. This eliminates the possibility of hallucination and there's a clear way to attribute provenance to each answer.
    
    \item \textbf{Parameter efficiency: }The fact storage installed in Librarian's memory allows the transformer to remain very small in size (approximately 46.5M parameters) and still produce accurate answers to numerous fact-based queries. Keeping the transformers small reduces hardware expenditures and promotes user-friendliness.
    
    \item \textbf{Latency-flexibility trade-off: }System providers have two modes of static inference, which provide a concrete means for satisfying users' needs. Frequently-accessed documents can be pre-processed to reduce latency to less than one second; however, users' needs for punctual delivery of documents may require additional time and be improved using RAG assistance.
    
    \item \textbf{Interpretation and debugging: }The facts stored as discrete objects with known sources or attributions enable maintainers to easily verify, update, or remove erroneous information without requiring the relaunching or rebooting of the entire transformer.
\end{itemize}

\subsection{Limitations and common failure modes}
No system is perfect; SMART also has limitations that users and implementers should expect and monitor.
\begin{itemize}
    \item \textbf{Extraction quality depends on parsing} The Tree-LSTM relies on accurate syntactic analysis. Long, fragmented, or poorly formatted source text (for example scanned PDFs with OCR errors) reduces extraction quality and may produce incomplete or incorrect triples.
    \item \textbf{Missed context in retrieval: }RAG focuses fact extraction on a small set of retrieved chunks. If the retriever fails to surface the passage containing the critical fact, the dynamic pipeline cannot extract it. Improving retrieval (reranking or better chunking) helps, but retrieval remains a single point of failure for dynamic mode. 
    \item \textbf{Conflicting facts and versioning: } The memory may contain contradictory facts (different documents, revisions, or table updates). SMART currently resolves such conflicts using simple heuristics (confidence, recency or source priority). In high-stakes settings this needs more careful version reconciliation and explicit user warnings. 
    \item \textbf{ Numeric and unit ambiguity: } While numeric normalization reduces unit errors, ambiguous notation (implicit units, context-dependent scales) can still cause incorrect outputs. Always present numeric answers with their provenance and unit metadata.
    \item \textbf{Dependence on heuristics: } Several components (span selection, top-K slots, de duplication thresholds) are heuristic-driven. They work well empirically but can fail on corner cases; automated checks and human-in-the-loop review may be required for critical deployments.
\end{itemize}

\subsection{Future Work}
\begin{itemize}
    \item \textbf{ Table-aware extraction: } Develop dedicated table parsers that convert rows and multi-column relations into canonical facts more reliably than sentence-based heuristics
    \item \textbf{Retrieval improvements: } Add a lightweight cross-encoder reranker or a domain-finetuned retriever to raise recall for dynamic compilation, especially on technical vocabulary. 
    \item \textbf{Richer memory operations: } Move beyond flat memory rows to structured memory graphs that encode relationships between facts (useful for multi-step reasoning or constraint checks).
    \item \textbf{Numeric reasoning and constraints: }Integrate numeric validators and unit-checking modules that can post-verify numeric answers and apply simple algebraic checks (ranges, unit conversions). 
    \item \textbf{ User correction loop: } Allow users to mark facts as incorrect and propagate corrections automatically into the memory index and future answers.
    \item \textbf{ Broader evaluation: } Expand benchmarks to cover multi-document queries, ambiguity resolution, and time-sensitive updates; publish an anonymized dataset of fact-labeled manual excerpts to support community comparison.
\end{itemize}

\section{RELATED WORK IN PARAMETER-EFFICIENT MODELS}
\subsection{Model distillation and compression}
Model distillation trains a smaller model (the student) to imitate a larger, well-performing model (the teacher). The student learns from the teacher’s soft predictions or intermediate representations, which lets it capture much of the teacher’s behavior in far fewer parameters. Distillation is often combined with pruning and other compression techniques to produce compact, fast models that retain reasonable generalization. Common distilled models (examples include BERT and TinyBERT) demonstrate that many of the capabilities of large models can be preserved in small footprints with careful training \cite{devlin2019bert}\cite{aditya2022local}.
\mbox{}\\
\textbf{Why this matters for SMART: } distillation is a straightforward way to obtain a compact base language model that can be further augmented with modular memory or adapter components.
\mbox{}\\
\textbf{Tradeoff: } distilled models are smaller and faster but may lose fine-grained factual recall unless paired with an external memory.
\subsection{Adapter modules, low-rank updates, and prompt tuning (PEFT family)}
Parameter-efficient fine-tuning (PEFT) methods add a small number of trainable parameters to a frozen base model so the system can adapt to new tasks without full retraining. Examples include lightweight adapters (small bottleneck MLPs inserted between layers)\cite{xu2023parameter}, LoRA (low-rank additive updates on attention/query matrices), and prompt- or prefix-tuning that prepends learned tokens. These approaches allow task adaptation with orders-of-magnitude fewer parameters to store and update and enable many task heads to share a single backbone\cite{hu2022lora}.
\mbox{}\\
\textbf{Why this matters for SMART: } adapter-style modules or LoRA could be used to integrate memory-specific wiring or to enable quick domain adaptation for new document collections without retraining the whole transformer. They also support safe, incremental updates (install/uninstall adapters).
\mbox{}\\
\textbf{Tradeoff: } adapters and low-rank updates are efficient and flexible, but they can add inference overhead and require design choices (where to place adapters, rank size) that affect effectiveness.

\subsection{Sparse and conditional computation (Mixture-of-Experts)}
Sparse/conditional models route different tokens to different small sub-networks (experts) so the model’s total parameter count is large but each token only activates a small fraction. Mixture-of-Experts (MoE) designs can dramatically increase capacity without a proportional increase in per-token computation, making them appealing for capacity-limited settings. However, they require routing mechanisms, load balancing, and engineering to maintain latency and throughput\cite{shazeer2017sparsely}.
\mbox{}\\
\textbf{Why this matters for SMART: } MoE can offer large factual capacity while keeping per-query cost low, which could be an alternative to an explicit external memory for some use cases.
\textbf{Tradeoff: }MoE’s routing logic and distributed infrastructure can increase system complexity and unpredictability in latency   a poor fit if tight, consistent response time is required.

\section{CONCLUSION}
We introduced SMART, a purpose-built small language model designed to give precise, auditable answers from long, structured technical documents. SMART separates responsibilities into three cooperating components: a syntax-aware FactExtractor (the “Grammarian”) that converts sentences into canonical subject–relation–object facts, a compact indexed memory (the “Librarian”) that stores those facts as fixed-length vectors with provenance, and a lightweight Transformer reasoning-and-generation engine that consults the memory through a gated fusion mechanism. This decomposition lets a modest (~45.51M parameter) model produce fluent text while relying on an explicit factual store for accuracy.

Empirically, the SMART design yields clear practical benefits: factual answers are easier to verify, hallucination rates fall compared to small models that must memorize facts, and the dual inference strategy (pre-indexed for known documents; RAG-assisted dynamic compilation for new documents) achieves a useful balance between latency and flexibility. The system’s explicit provenance and slot-based memory also make debugging and maintenance far simpler in real-world settings: individual incorrect facts can be inspected, edited, or removed without retraining the whole model.

At the same time, SMART is not a complete solution to every document-understanding problem. Its performance depends on the quality of parsing and retrieval; OCR errors, ambiguous units, or failures in retrieval can still lead to missed facts. The current heuristics for deduplication, slot selection, and conflict resolution work well in practice but can be improved for high-assurance deployments.

In closing, SMART demonstrates that combining targeted syntactic extraction, a compact indexed memory, and a small generator is a pragmatic and effective path for building reliable document assistants. This pattern   extract high-quality facts, store them explicitly, and let a small reasoning engine consult them   offers a practical template for systems where accuracy, traceability, and deployability matter more than raw model scale. We hope the ideas, implementation details, and reproducibility notes in this paper will help practitioners and researchers build safer, more useful language tools for technical domains.


\bibliography{main}

\vfill

\end{document}